\documentclass{article}

\usepackage{microtype}
\usepackage{graphicx}
\usepackage{subfigure}
\usepackage{booktabs}

\usepackage{graphicx}
\usepackage{glossaries}
\usepackage[export]{adjustbox}
\usepackage{enumitem}
\usepackage{amssymb}
\usepackage{bm}

\usepackage{hyperref}

\usepackage[accepted]{icml2018}

\icmltitlerunning{ChatPainter: Improving Text to Image Generation using Dialogue}

\begin{document}

\newacronym{cnn}{CNN}{Convolutional Neural Network}
\newacronym{coco}{MS COCO}{Microsoft Common Objects in Context}
\newacronym{gan}{GAN}{Generative Adversial Network}
\newacronym{rnn}{RNN}{Recurrent Neural Network}
\newacronym{vae}{VAE}{Variational Autoencoder}
\newacronym{visdial}{VisDial}{Visual Dialog dataset}
\newacronym{amt}{AMT}{Amazon Mechanical Turk}
\newacronym{ca}{CA}{Conditioning Augmentation}
\newacronym{dcgan}{DC-GAN}{Deep Convolutional Generative Adversarial Network}
\newacronym{wgan}{WGAN}{Wasserstein GAN}

\twocolumn[
\icmltitle{ChatPainter: Improving Text to Image Generation using Dialogue}

\icmlsetsymbol{equal}{*}

\begin{icmlauthorlist}
\icmlauthor{Shikhar Sharma}{msr}
\icmlauthor{Dendi Suhubdy}{udem,mila}
\icmlauthor{Vincent Michalski}{udem,mila,msr}
\icmlauthor{Samira Ebrahimi Kahou}{msr}
\icmlauthor{Yoshua Bengio}{udem,mila}
\end{icmlauthorlist}

\icmlaffiliation{msr}{Microsoft Research, Montr{\'e}al, Canada}
\icmlaffiliation{udem}{Universit{\'e} de Montr{\'e}al, Montr{\'e}al, Canada}
\icmlaffiliation{mila}{Montreal Institute for Learning Algorithms, Montr{\'e}al, Canada}

\icmlcorrespondingauthor{Shikhar Sharma}{shikhar.sharma@microsoft.com}

\icmlkeywords{text to image generation, dialogue to image generation, image generation, gan, coco}

\vskip 0.3in
]

\printAffiliationsAndNotice{}

\begin{abstract}
Synthesizing realistic images from text descriptions on a dataset like \gls{coco}, where each image can contain several objects, is a challenging task. Prior work has used text captions to generate images. However, captions might not be informative enough to capture the entire image and insufficient for the model to be able to understand which objects in the images correspond to which words in the captions. We show that adding a dialogue that further describes the scene leads to significant improvement in the inception score and in the quality of generated images on the \gls{coco} dataset.
\end{abstract}

\section{Introduction}
\label{sec:introduction}
Automatic generation of realistic images from text descriptions has numerous potential applications, for instance in image editing, in video games, or for accessibility. Spurred by the recent successes of \glspl{vae}~\cite{ICLR2014_vae} and \glspl{gan}~\cite{NIPS2014_gan, NIPS2015_lapgan, ICLR2016_dcgan}, there has been a lot of recent work and interest in the research community on image generation from text captions~\cite{ICLR2016_elman, ICML2016_reed, Zhang_2017_ICCV, attngan}.

\begin{figure}[ht]
\vskip 0.2in
\begin{center}
\centerline{
\begin{minipage}[b]{0.23\textwidth}
\small
This flower has overlapping pink pointed petals surrounding a ring of short yellow filaments
\end{minipage}
\qquad
\includegraphics[width=0.12\textwidth,valign=b]{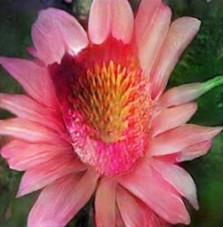}
}
\caption{Caption and corresponding generated image from the StackGAN model \cite{Zhang_2017_ICCV}. Image reproduced with permission from authors.}
\label{fig:flower}
\end{center}
\vskip -0.2in
\end{figure}

\begin{figure}[ht]
\vskip 0.2in
\begin{center}
\centerline{
\hfill
\subfigure[A flock of birds flying in a blue sky.]{\makebox[35mm][c]{\includegraphics[width=0.22\textwidth]{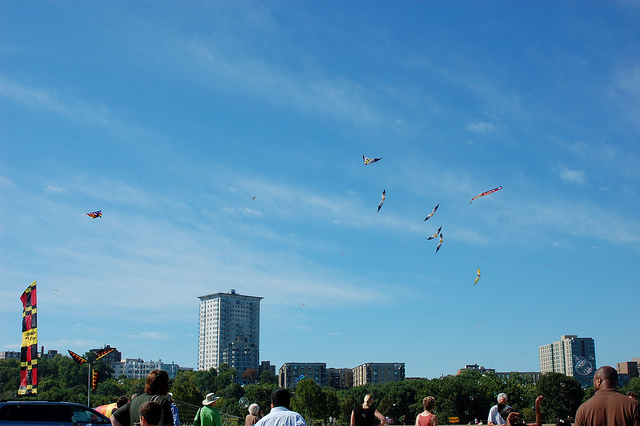}}}
\hfill
\subfigure[A flock of birds flying in an overcast sky]{\makebox[35mm][c]{\includegraphics[width=0.22\textwidth]{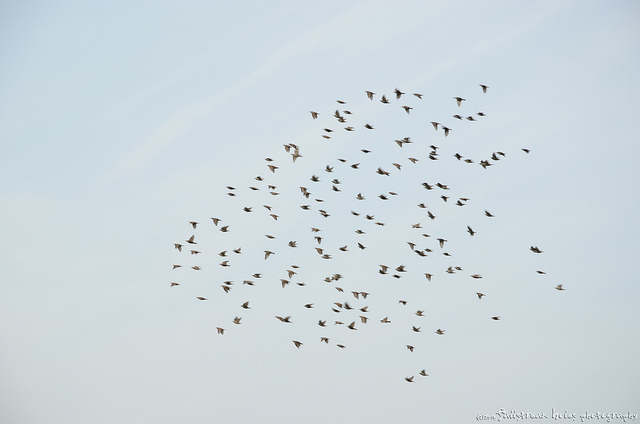}}}
\hfill
}
\caption{Two very different looking images can have similar captions in the COCO dataset and the captions also might not describe the image fully.}
\label{fig:coco-similar}
\end{center}
\vskip -0.2in
\end{figure}

Current state-of-the-art models are capable of generating realistic images on datasets of birds, flowers, room interiors, faces etc., but don't do very well on datasets like \gls{coco}~\cite{lin2014microsoft} which contain several objects within a single image and where subjects are not always centred in the image. A caption for an image of a flower can usually describe most of the relevant details of the flower (see Figure~\ref{fig:flower}). However, for the \gls{coco} dataset a caption might not contain all the relevant details about the foreground and the background. As can be seen from Figure~\ref{fig:coco-similar}, it is possible for two very similar \gls{coco} captions to correspond to very different images. Due to this complexity, a caption can be considered a noisy descriptor and due to the limited amount of image-caption paired data, the model might not always be able to understand which objects in the image correspond to which words in the caption. Previous work in the literature has found that conditioning on auxiliary data such as category labels~\cite{cgan}, or object location and scale~\cite{NIPS2016_reed} helps in improving the quality of generated images and in making them more interpretable.

Sketch artists typically have a back and forth conversation with witnesses when they have to draw a person's sketch, where the artist asks for more details and draws the sketch while the witnesses provide requested details and feedback on the current state of the sketch. 
We hypothesize that conditioning on a similar conversation about a scene in addition to a caption would significantly improve the generated image's quality and we explore this idea in this paper.
For this, we pair captions provided with the \gls{coco} dataset with dialogues from the \gls{visdial} \cite{das2016visual}. These dialogues were collected using a chat interface pairing two workers on \gls{amt}. One of them was assigned the role of an `answerer', who could see an \gls{coco} image together with its caption and had to answer questions about that image. The other was assigned the role of the `questioner' and could see only the image's caption. 
The questioner had to ask questions to be able to imagine the scene more clearly.
Similar to a dialogue between a sketch artist who gradually refines an image and a witness describing a person, the \gls{visdial} dialogue turns iteratively add fine-grained details to a (mental) image. 

In this paper,
\begin{itemize}
    \item We use \gls{visdial} dialogues along with \gls{coco} captions to generate images. We show that this results in the generation of better quality images. 
    \item We provide results indicating that our model obtains a higher inception score than the baseline StackGAN model which uses only captions.
\end{itemize}
Though we just demonstrate improvements over the StackGAN~\cite{Zhang_2017_ICCV} model in this paper, this additional dialogue module can be added to any caption-to-image-generation model and is an orthogonal contribution.

\begin{figure*}[ht]
\vskip 0.2in
\begin{center}
\centerline{
\subfigure[Stage-I model]{\includegraphics[width=\textwidth]{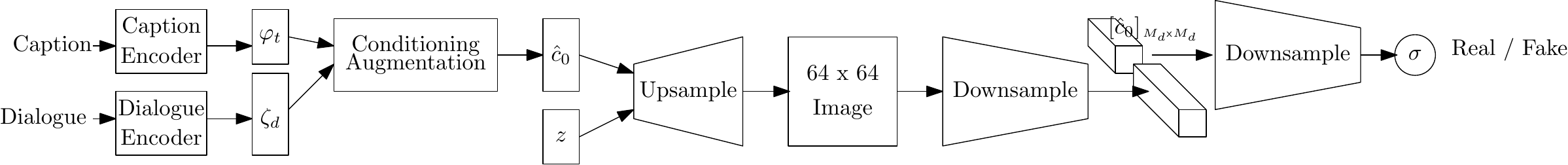}}
}
\centerline{
\subfigure[Stage-II model]{\includegraphics[width=\textwidth]{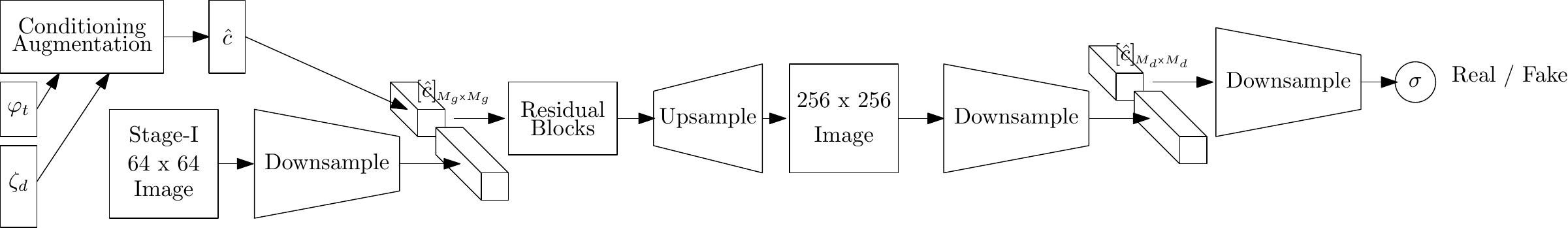}}
}
\caption{ChatPainter: (a) Stage-I of the model generates a $64\times64$ image conditioned on a caption and the corresponding dialogue. (b) Stage-II of the model generates a $256\times256$ image conditioned on Stage-I's $64\times64$ generated image and the caption and corresponding dialogue.}
\label{fig:model}
\end{center}
\vskip -0.2in
\end{figure*}

\section{Related Work}
\label{sec:related-work}
In the past, variationally trained models have often been used for image generation. A significant drawback of these models has been that they tend to generate blurry images. Among latent variable based variationally trained models,  \citet{NIPS2016_iaf} proposed inverse autoregressive flows, where they inverted the sequential data generation of an autoregressive model, which helped in parallelizing computation. They presented results on MNIST \cite{lecun1998gradient} and CIFAR-10~\cite{krizhevsky2009cifar} datasets. Recently, \citet{deepresidualvae} added residual blocks \cite{he2016deep} and skip connections to their decoder and generated images in multiple stages. The initial components produce a coarse image and the final components refine the previously generated image. Their deep residual \gls{vae} was able to generate sharper images on MNIST and CelebA~\cite{Liu_2015_ICCV_celeba} compared to previous \glspl{vae}. Among tractable likelihood models, \citet{pixelrnn} proposed the PixelRNN model which predicts pixels of an image sequentially along the rows or along the diagonal using fast two-dimensional recurrent layers. They achieved significantly improved log-likelihood scores on the MNIST, CIFAR-10 and ImageNet~\cite{imagenet} datasets. \citet{makhzani_aae} proposed an adversarial autoencoder model where they use \glspl{gan} to perform variational inference and achieve competitive performance on both generative and semi-supervised classification tasks.

\glspl{gan} have received attention recently because they produce sharper images \cite{NIPS2014_gan, NIPS2015_lapgan, ICLR2016_dcgan} compared to other generative models. They however suffer from several issues such as `mode collapse' (i.e. the generator learns to generate samples from only a few modes of the distribution), lack of variation among generated images, and instabilities in the training procedure. Training \glspl{gan} has generally required careful design of the model architecture and a balance between optimization of the generator and the discriminator. 
\citet{wgan} proposed minimizing an approximation of the Earth Mover (Wasserstein) distance between the real and generated distributions. Their model, \gls{wgan}, is much more stable than previous approaches and reduces most of the aforementioned issues affecting \glspl{gan}. Additionally, the reduction of the critic's (the discriminator is called the `critic' in this work) loss correlates with better sample quality which is a desirable property. \citet{wgan-gp} further improved upon these issues by removing weight clipping from the \gls{wgan} and instead adding a gradient penalty (WGAN-GP) for the gradient of the critic. Recently, \citet{karras2018progressive} have produced high-quality high resolution images ($1024\times1024$) by progressively growing both the generator and the discriminator layer-by-layer and by using the WGAN-GP loss. Apart from faster training time, they found that by adding layers progressively, training stability is improved significantly. Among recent efforts to stabilize the training of \glspl{gan} via noise-induced regularization, \citet{NIPS2017_6797} reduced several failure modes of \glspl{gan} by penalizing a weighted gradient-norm of the discriminator. They observed stability improvements and better generalization performance. \citet{miyato2018spectral} proposed a spectral weight normalization technique for \glspl{gan} in which they control the Lipschitz constant of the discriminator function resulting in global regularization of the discriminator. Gradient analysis of the spectrally normalized weights shows that their technique prevents layers from becoming sensitive in a single direction. This approach yields more complexity and variation in generated samples compared to previous weight normalization methods.

Apart from generating images directly from noise with \glspl{gan}, there has been recent work on conditioning the generator or discriminator or both on additional information. \citet{cgan} introduced the idea of conditioning both the generator and discriminator on extra information such as class labels. They ran experiments on both unimodal image data and multi-modal image\nobreakdash-metadata\nobreakdash-annotations data. Their experiments resulted in better Parzen window-based log-likelihood estimates for MNIST compared to unconditioned \glspl{gan}. On the MIR Flickr 25\,000 dataset~\cite{huiskes08}, they generated metadata tags conditioned on images. \citet{acgan} conditioned their generator on both noise and class labels and their discriminator additionally classified images into classes. By explicitly making the generator and discriminator aware of class labels, they were able to generate better quality images on larger multi-class datasets and higher image variability compared to previous work. Their experiments also indicated that generating higher resolution images yielded higher discriminability.

A lot of work has been done in recent years to generate \gls{coco} images from captions. \citet{ICLR2016_elman} used a conditional DRAW~\cite{draw} model with soft attention over the words of the caption to generate images on the \gls{coco} dataset and then sharpened them with an adversarial network. However, their generated images were low resolution ($32\times32$) and generated blob-like objects in most cases. Building upon other work in \glspl{gan}, \citet{ppgn} introduced a prior on the latent code used by their generator. 
They ran an optimization procedure to find the latent code which the generator takes as its input. This procedure maximized the activations of an image captioning network run over the generated image. This method produced high-quality and diverse images at high resolution ($227\times227$). \citet{ICML2016_reed} trained a \gls{dcgan} with the generator and discriminator both conditioned on features from a character-level convolutional \gls{rnn} encoder over the captions to generate visually-plausible $64\times64$ images. \citet{tac-gan} additionally trained their discriminator to classify images similar to \citet{acgan} and generated $128\times128$ images. \citet{Zhang_2017_ICCV} also conditioned their generator and discriminator on caption encodings but their StackGAN model generates images in multiple stages -- stage-I generates a coarse low resolution ($64\times64$) image and stage-II generates the final high resolution ($256\times256$) image. This stacking of models resulted in generation of highly photo-realistic images, at higher resolutions compared to previous work, on datasets of flowers~\cite{flowers} and birds~\cite{WahCUB_200_2011_birds} and many good looking images on the \gls{coco} dataset as well. However, they did not train their two stages in an end-to-end fashion. Stage-I was trained to completion first and then Stage-II was trained. Very recently, \citet{attngan} increased the number of stages, trained them in an end-to-end fashion, added an attention mechanism over the captions, as well as added a novel attentional multimodal similarity model to guide the training loss, which resulted in significantly increased performance and the state-of-the-art inception score~\cite{NIPS2016_6125_inception} of $25.89$ on the \gls{coco} dataset. \citet{honglak2018} first generated a semantic layout map of the objects in the image and then conditioned on the map and the caption to generate semantically meaningful $128\times128$ images.

\begin{table*}[ht]
    \begin{center}
	\caption{An example of the input data, the corresponding dataset image, and the image generated by our best ChatPainter model.}
	\label{table:caption-dialogue-image}
		\begin{tabular}{lcc} \hline
			\multicolumn{1}{l}{\bf Input} &\multicolumn{1}{c}{\bf Dataset image} &\multicolumn{1}{c}{\bf Generated image}\\ \hline
			\begin{minipage}{0.6\textwidth}
			Caption: adult woman with yellow surfboard standing in water.\\
			Q: is the woman standing on the board? \hfill A: no she is beside it.\\
            Q: how much of her is in the water? \hfill A: up to her midsection.\\
            Q: what color is the board? \hfill A: yellow.\\
            Q: is she wearing sunglasses? \hfill A: no.\\
            Q: what about a wetsuit? \hfill A: no she has on a bikini top.\\
            Q: what color is the top? \hfill A: orange and white.\\
            Q: can you see any other surfers? \hfill A: no.\\
            Q: is it sunny? \hfill A: the sky isn't visible but it appears to be a nice day.\\
            Q: can you see any palm trees? \hfill A: no.\\
            Q: what about mountains? \hfill A: no.\end{minipage} & \includegraphics[valign=m, width=0.15\textwidth]{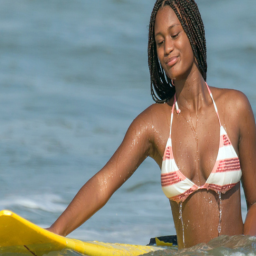} & \includegraphics[valign=m, width=0.15\textwidth]{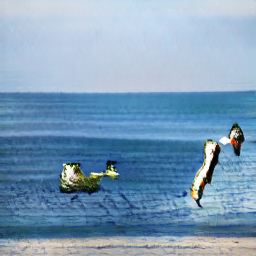} \\ \hline
		\end{tabular}
	\end{center}
\end{table*}

\section{Data}
\label{sec:data}
In our experiments, we used images and their captions from the \gls{coco} \citep{lin2014microsoft} dataset. \gls{coco} covers 91 categories of objects, grouped into 11 super-categories of objects such as \emph{person and accessory}, \emph{animal}, \emph{vehicle}, etc. We use the `2014 Train' set as our training set and the `2014 Val' set as our test set. The train set consists of $\sim 80K$ images and includes five captions for each image. The test set consists of $\sim 40K$ images along with their captions.

We obtain dialogues for these images from  the \gls{visdial}~\cite{das2016visual} dataset. \gls{visdial} consists of 10 question-answer conversation turns per dialogue and has one dialogue for each of the \gls{coco} images. \gls{visdial} was collected by pairing two crowd-workers and having them talk about an image as described in Section~\ref{sec:introduction}. Hence, we have $\sim 80K$ dialogues for the training set and $\sim 40K$ for the test set.

\section{Model}
\label{sec:model}
We build upon the StackGAN model introduced by \citet{Zhang_2017_ICCV}. StackGAN generates an image in two stages where Stage-I generates a coarse $64\times64$ image and Stage-II generates a refined $256\times256$ image. We try to use the same notation everywhere as used in the original StackGAN paper. Our model ChatPainter's architecture is shown in Figure~\ref{fig:model} and described below.

We generate caption embedding $\varphi_t$ by encoding the captions with a pre-trained encoder\footnote{\scriptsize \url{https://github.com/reedscot/icml2016}}~\cite{ICML2016_reed}. We generate dialogue embeddings $\zeta_d$ by two methods:
\begin{itemize}[noitemsep,topsep=0pt]
    \item \textbf{Non-recurrent encoder} We collapse the entire dialogue into a single string and encode it with a pre-trained Skip-Thought~\cite{NIPS2015_skipthought} encoder~\footnote{\scriptsize \url{https://github.com/ryankiros/skip-thoughts}}.
    \item \textbf{Recurrent encoder} We generate Skip-Thought vectors for each turn of the dialogue and then encode them with a bidirectional LSTM-RNN~\cite{graves2005framewise,hochreiter1997long}.
\end{itemize}

We then concatenate the caption and dialogue embeddings and this is passed as input to the \gls{ca} module. The \gls{ca} module was introduced by \citet{Zhang_2017_ICCV} to produce latent variable inputs for the generator from the embeddings. They also proposed a regularization term to encourage smoothness over the conditioning manifold which we adapt for our additional dialogue embeddings:
\begin{align}
    D_{KL}(\mathcal{N}(\mu(\varphi_t, \zeta_d), \text{diag}(\sigma(\varphi_t, \zeta_d))) || \mathcal{N}(0, I)),
\end{align}
where $D_{KL}$ is the Kullback-Leibler divergence. In the \gls{ca} module, a fully connected layer is applied over the input that generates $\mu$ and $\sigma$ which are both $N_g$ dimensional. The module samples $\epsilon$ from $\mathcal{N}(0, I)$. Finally, the conditioning variables $\hat{c}$ are computed as
\begin{align}
    \hat{c} &= \mu + \sigma \odot \epsilon,
\end{align}
where $\odot$ is the element-wise multiplication operator. Thus, the conditioning variables $\hat{c}$ are effectively samples from $\mathcal{N}(\mu(\varphi_t, \zeta_d), \text{diag}(\sigma(\varphi_t, \zeta_d)))$.

\begin{figure*}[h]
    \subfiguretopcapfalse
    \subfigure{\includegraphics[width=0.135\textwidth]{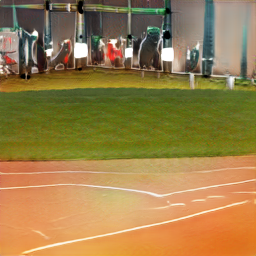}}
    \hfill
    \subfigure{\includegraphics[width=0.135\textwidth]{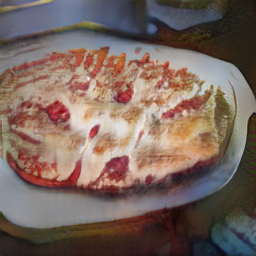}}
    \hfill
    \subfigure{\includegraphics[width=0.135\textwidth]{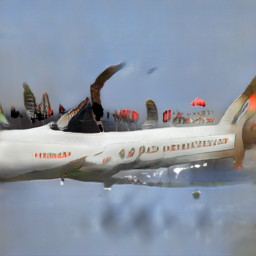}}
    \hfill
    \subfigure{\includegraphics[width=0.135\textwidth]{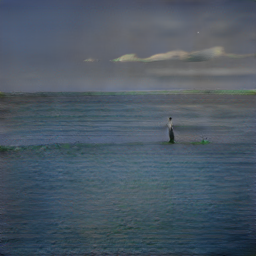}}
    \hfill
    \subfigure{\includegraphics[width=0.135\textwidth]{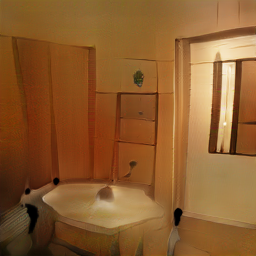}}
    \hfill
    \subfigure{\includegraphics[width=0.135\textwidth]{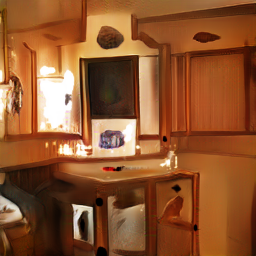}}
    \hfill
    \subfigure{\includegraphics[width=0.135\textwidth]{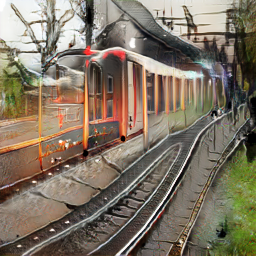}}
    \hfill
    \subfigure{\includegraphics[width=0.135\textwidth]{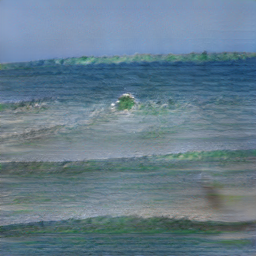}}
    \hfill
    \subfigure{\includegraphics[width=0.135\textwidth]{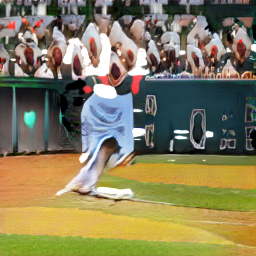}}
    \hfill
    \subfigure{\includegraphics[width=0.135\textwidth]{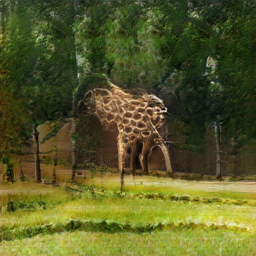}}
    \hfill
    \subfigure{\includegraphics[width=0.135\textwidth]{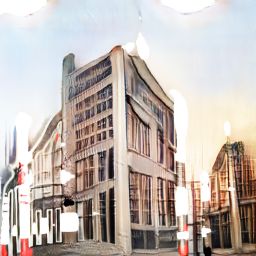}}
    \hfill
    \subfigure{\includegraphics[width=0.135\textwidth]{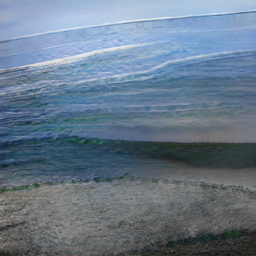}}
    \hfill
    \subfigure{\includegraphics[width=0.135\textwidth]{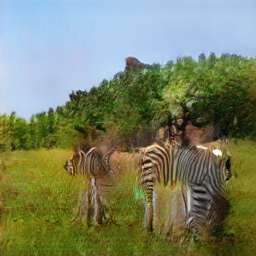}}
    \hfill
    \subfigure{\includegraphics[width=0.135\textwidth]{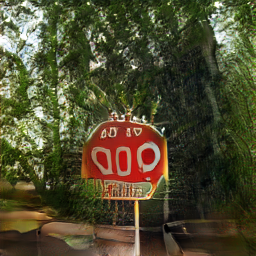}}
    \hfill
    \subfigure{\includegraphics[width=0.135\textwidth]{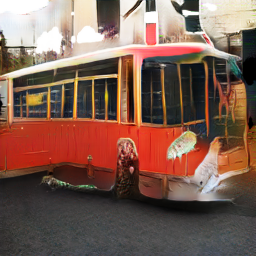}}
    \hfill
    \subfigure{\includegraphics[width=0.135\textwidth]{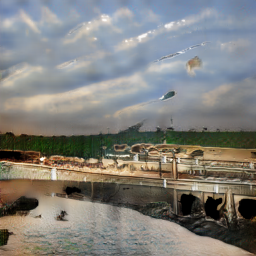}}
    \hfill
    \subfigure{\includegraphics[width=0.135\textwidth]{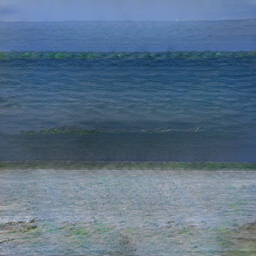}}
    \hfill
    \subfigure{\includegraphics[width=0.135\textwidth]{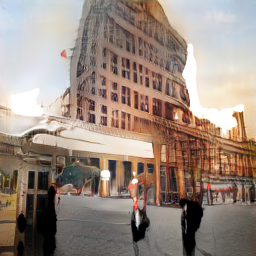}}
    \hfill
    \subfigure{\includegraphics[width=0.135\textwidth]{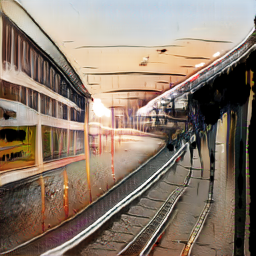}}
    \hfill
    \subfigure{\includegraphics[width=0.135\textwidth]{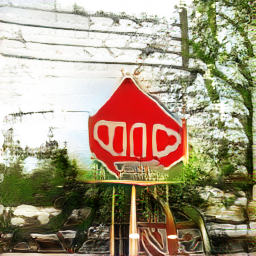}}
    \hfill
    \subfigure{\includegraphics[width=0.135\textwidth]{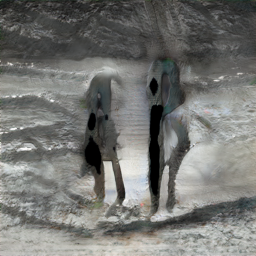}}
    \caption{Example $256\times256$ images generated by our non-recurrent encoder ChatPainter model on the \gls{coco} test set. Best viewed in color. Images are cherry-picked from a larger random sample.}
    \label{fig:genimgsnonrec}
\end{figure*}

\begin{figure*}[!h]
    \subfiguretopcapfalse
    \subfigure{\includegraphics[width=0.135\textwidth]{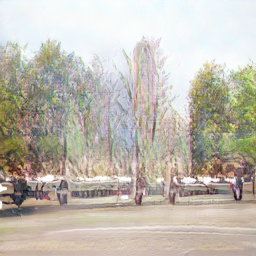}}
    \hfill
    \subfigure{\includegraphics[width=0.135\textwidth]{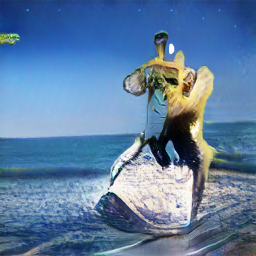}}
    \hfill
    \subfigure{\includegraphics[width=0.135\textwidth]{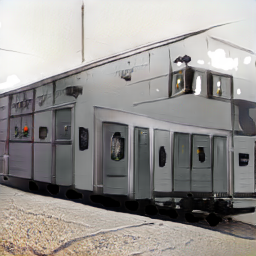}}
    \hfill
    \subfigure{\includegraphics[width=0.135\textwidth]{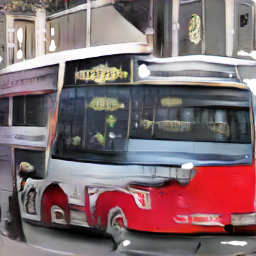}}
    \hfill
    \subfigure{\includegraphics[width=0.135\textwidth]{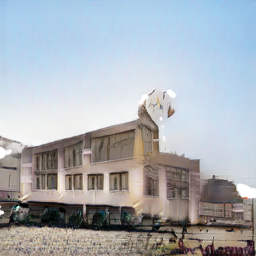}}
    \hfill
    \subfigure{\includegraphics[width=0.135\textwidth]{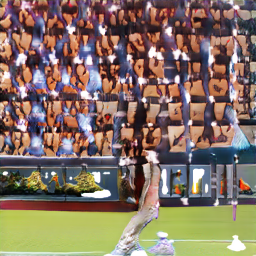}}
    \hfill
    \subfigure{\includegraphics[width=0.135\textwidth]{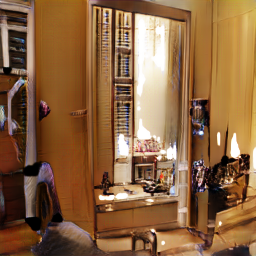}}
    \hfill
    \subfigure{\includegraphics[width=0.135\textwidth]{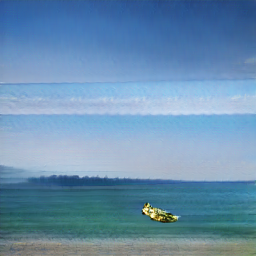}}
    \hfill
    \subfigure{\includegraphics[width=0.135\textwidth]{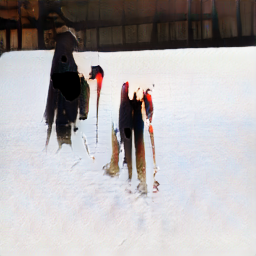}}
    \hfill
    \subfigure{\includegraphics[width=0.135\textwidth]{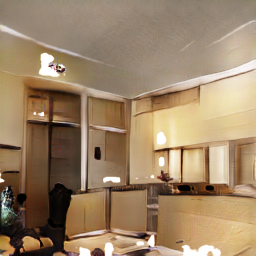}}
    \hfill
    \subfigure{\includegraphics[width=0.135\textwidth]{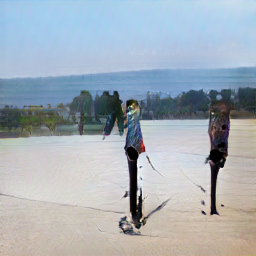}}
    \hfill
    \subfigure{\includegraphics[width=0.135\textwidth]{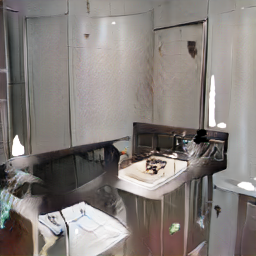}}
    \hfill
    \subfigure{\includegraphics[width=0.135\textwidth]{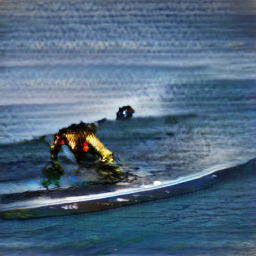}}
    \hfill
    \subfigure{\includegraphics[width=0.135\textwidth]{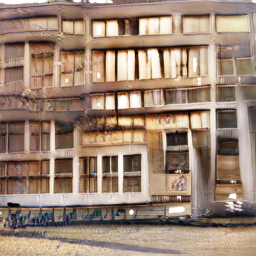}}
    \hfill
    \subfigure{\includegraphics[width=0.135\textwidth]{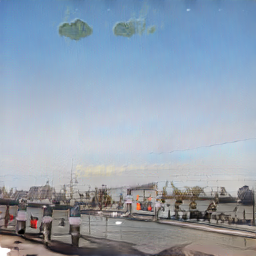}}
    \hfill
    \subfigure{\includegraphics[width=0.135\textwidth]{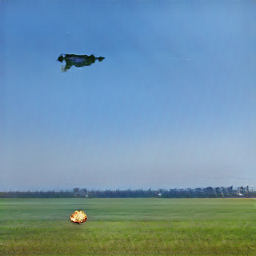}}
    \hfill
    \subfigure{\includegraphics[width=0.135\textwidth]{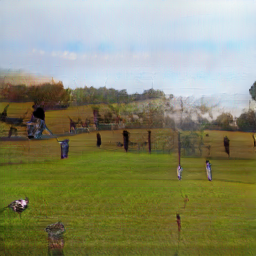}}
    \hfill
    \subfigure{\includegraphics[width=0.135\textwidth]{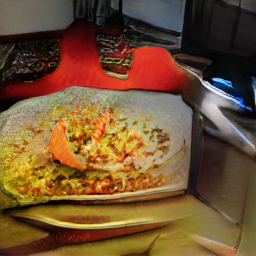}}
    \hfill
    \subfigure{\includegraphics[width=0.135\textwidth]{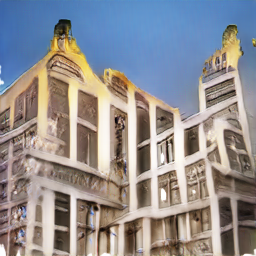}}
    \hfill
    \subfigure{\includegraphics[width=0.135\textwidth]{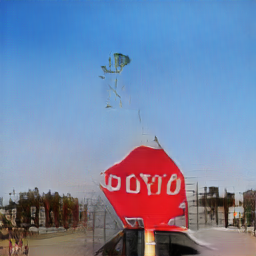}}
    \hfill
    \subfigure{\includegraphics[width=0.135\textwidth]{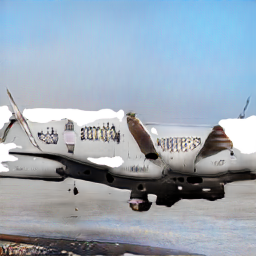}}
    \caption{Example $256\times256$ images generated by our recurrent encoder ChatPainter model on the \gls{coco} test set. Best viewed in color. Images are cherry-picked from a larger random sample.}
    \label{fig:genimgsrec}
\end{figure*}

\begin{figure*}[!h]
    \includegraphics[width=\textwidth]{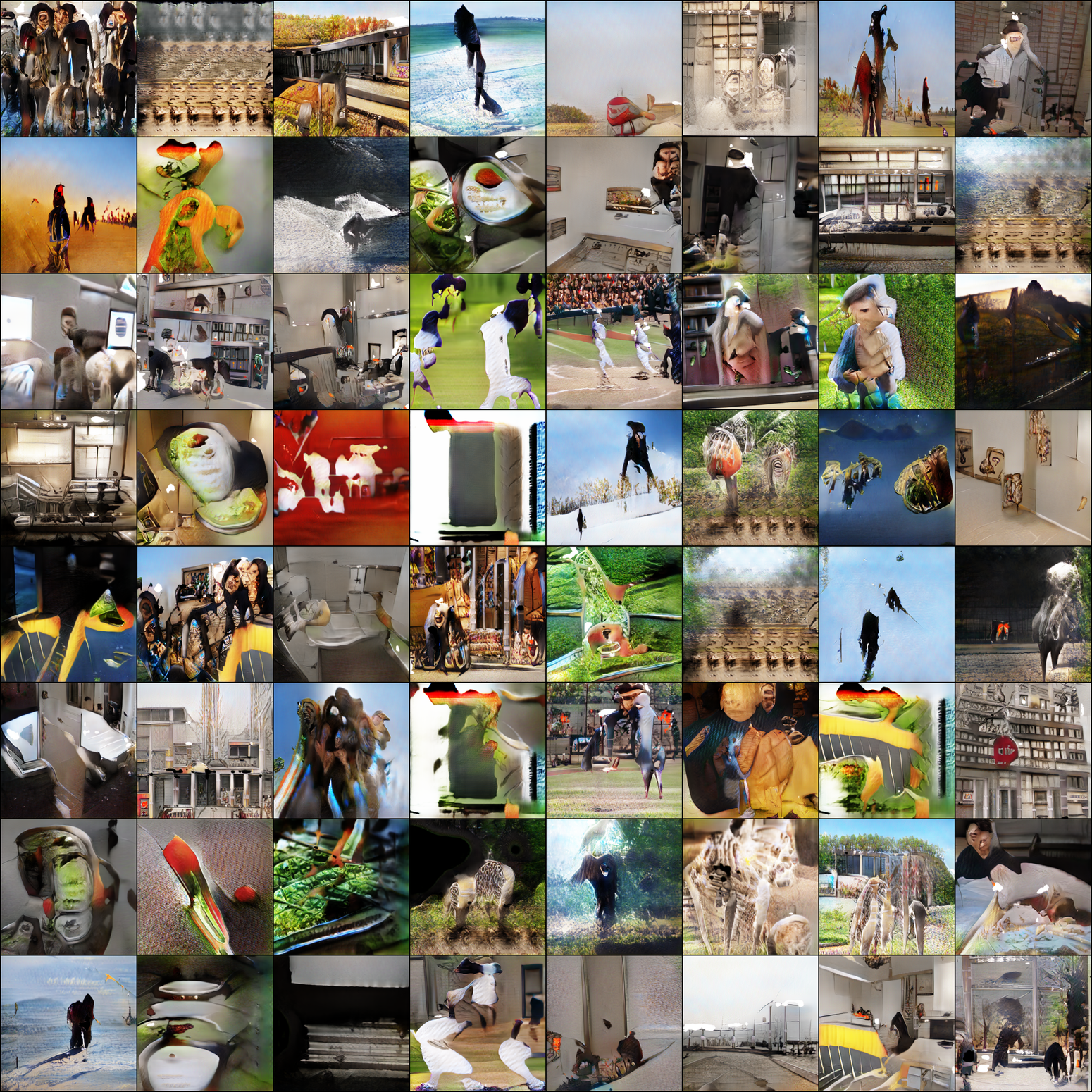}
    \caption{Example $256\times256$ images generated by our recurrent encoder ChatPainter model on the \gls{coco} test set. Best viewed in color. Images have been selected randomly.}
    \label{fig:genimgsgridrec}
\end{figure*}

\subsection{Stage-I}
The conditioning variables for Stage-I, $\hat{c}_0$, are concatenated with $N_z$-dimensional noise, $z$, drawn from a random normal distribution, $p_z$. The Stage-I generator upsamples this input representation to a $W_0 \times H_0$ image. This Stage-I image is expected to be blurry and a rough version of the final one. The discriminator downsamples this image to $M_d \times M_d \times N_{di}$. $\hat{c}_0$ is then spatially replicated to $M_d \times M_d \times N_d$ and concatenated with the downsampled representation. This is further downsampled to a scalar value between 0 and 1. The model is trained by alternating between maximizing $\mathcal{L}_{D_0}$ and minimizing $\mathcal{L}_{G_0}$:
\begin{align}
\begin{split}
\mathcal{L}_{D_0} &= \mathbb{E}_{(I_0,t,d)\sim p_{data}}[\log D_0(I_0,\varphi_t,\zeta_d)] + \\
& \mathbb{E}_{z\sim p_z, (t,d)\sim p_{data}}[\log(1-D_0(G_0(z,\hat{c}_0),\varphi_t,\zeta_d))],
\end{split}\\
\begin{split}
\mathcal{L}_{G_0} &= \mathbb{E}_{z\sim p_z, (t,d)\sim p_{data}}[\log(1-D_0(G_0(z,\hat{c}_0),\varphi_t,\zeta_d))] \\
& + \lambda D_{KL}(\mathcal{N}(\mu(\varphi_t, \zeta_d), \text{diag}(\sigma(\varphi_t, \zeta_d))) || \mathcal{N}(0, I)),
\end{split}
\end{align}
where $I_0$ is the real image, $t$ is the text caption, $d$ is the dialogue, $p_{data}$ is the true data distribution, $\lambda$ is the regularization coefficient, $G_0$ is the Stage-I generator, and $D_0$ is the Stage-I discriminator.

In our experiments, $N_z = 100$, $W_0 = 64$, $H_0 = 64$, $M_d = 4$, $N_{di} = 512$, $N_d = 128$, and $\lambda = 2$ -- same as that in the StackGAN model.

\subsection{Stage-II}
The Stage-II generator, $G$, first downsamples generated stage-I images to $M_g \times M_g \times N_{gi}$. The conditioning variables for Stage-II, $\hat{c}$, are generated and then spatially replicated to $M_g \times M_g \times N_g$ and finally concatenated to the downsampled image representation. For Stage-II training, in case of the recurrent dialogue encoder, the RNN weights are copied from Stage-I and kept fixed. The concatenated input is passed through a series of residual blocks and is then upsampled to a $W \times D$ image. The Stage-II discriminator, $D$, downsamples the input image to $M_d \times M_d \times N_{di}$. $\hat{c}$ is then spatially replicated to $M_d \times M_d \times N_d$ and concatenated with the downsampled representation which is further downsampled to a scalar value between 0 and 1. The Stage-II model is trained by alternating between maximizing $\mathcal{L}_D$ and minimizing $\mathcal{L}_G$:
\begin{align}
\begin{split}
\mathcal{L}_{D} &= \mathbb{E}_{(I,t,d)\sim p_{data}}[\log D(I,\varphi_t,\zeta_d)] + \\
& \mathbb{E}_{s_0\sim p_{G_0}, (t,d)\sim p_{data}}[\log(1-D(G(s_0,\hat{c}),\varphi_t,\zeta_d))],
\end{split}\\
\begin{split}
\mathcal{L}_{G} &= \mathbb{E}_{s_0\sim p_{G_0}, (t,d)\sim p_{data}}[\log(1-D(G(s_0,\hat{c}),\varphi_t,\zeta_d))] \\
& + \lambda D_{KL}(\mathcal{N}(\mu(\varphi_t, \zeta_d), \text{diag}(\sigma(\varphi_t, \zeta_d))) || \mathcal{N}(0, I)),
\end{split}
\end{align}
where $I$ is the real image, and $s_0$ is the image generated from Stage-I.

In our experiments, $M_g = 16$, $N_{gi} = 512$, $N_g = 128$, $W = 256$, $D = 256$, $N_{di} = 512$, $N_d = 128$, and $\lambda = 2$ -- same as that in the StackGAN model.

The architecture of the upsample, downsample and residual blocks, as shown in Figure~\ref{fig:model} and as mentioned above in the model details, is kept the same as that of the original StackGAN.

\subsection{Training details}
Similar to StackGAN, we use a matching-aware discriminator \cite{ICML2016_reed}, that is trained using ``real'' pairs consisting of a real image together with matching caption and dialogue, and ``fake'' pairs that consist either of a real image together with another images's caption and dialogue or a generated image with the corresponding caption and dialogue. We train both stages for $800$ epochs using the Adam optimizer \cite{kingma2014adam}. The initial learning rate for all experiments is $0.0002$. We decay the learning rate to half of its previous value after every $50$ epochs. For Stage-I, we use a batch size of $384$ and for Stage-II, we use a batch size of $64$. In case of the recurrent dialogue encoder, the hidden dimension of the RNN is set to $1024$. The implementation is based on \textit{PyTorch}~\cite{paszke2017automatic} and we trained the models on a machine with 4 NVIDIA Tesla P40s.

\setcounter{footnote}{1}
\begin{table}[t]
	\caption{Inception scores for generated images on the \gls{coco} test set\footnotemark.}
	\label{table:inception-score}
	\begin{center}
		\begin{tabular}{|r|r|}\hline
			\multicolumn{1}{|r|}{\bf Model}  &\multicolumn{1}{r|}{\bf Inception Score}
			\\ \hline
			\citet{ICML2016_reed}            & $7.88 \pm 0.07$ \\ \hline
			StackGAN~\cite{Zhang_2017_ICCV}  & $8.45 \pm 0.03$ \\
			ChatPainter (non-recurrent)      & $9.43 \pm 0.04$ \\
			ChatPainter (recurrent)          & $\mathbf{9.74 \pm 0.02}$ \\ \hline
			\citet{honglak2018}              & $11.46 \pm 0.09$ \\
			AttnGAN~\cite{attngan}           & $\mathbf{25.89 \pm 0.47}$ \\ \hline
		\end{tabular}
	\end{center}
\end{table}

\section{Results}
Table~\ref{table:caption-dialogue-image} shows the corresponding caption, dialogue inputs, and the test set image for an image generated by our best ChatPainter model. We present some of the more realistic images generated by our non-recurrent encoder ChatPainter in Figure~\ref{fig:genimgsnonrec}, and by our recurrent encoder ChatPainter in Figure~\ref{fig:genimgsrec}. For fairness of comparison, we also present a random sample of the images generated by our recurrent encoder ChatPainter on the \gls{coco} dataset in Figure~\ref{fig:genimgsgridrec}. As seen from these figures, the model is able to generate close-to-realistic images for some of the caption and dialogue inputs though not very realistic ones for most.

We report inception scores on the images generated from our models in Table~\ref{table:inception-score} and compare with other recent models\footnotetext{The two best-performing methods were released while writing this manuscript and we will evaluate the effect of our scheme using these methods as base architecture in future work.}. For computing inception score, we use the Inception v3 model pretrained on ImageNet available with \textit{PyTorch}. We then generate images for the 40k test set and use 10 random splits of 30k images each. We report the mean and standard deviation across these splits. We see that the ChatPainter model, which is conditioned on additional dialogue information, gets higher inception score than the StackGAN model just conditioned on captions. Also, the recurrent version of ChatPainter gets higher inception score than the non-recurrent version. This is likely due to it learning better encoding of the dialogues as the Skip-Thought encoder isn't trained with very long sentences, which is the case in the non-recurrent version where we collapse the dialogue in a single string.

\section{Discussion and Future Work}
In this paper, apart from conditioning on image captions, we additionally conditioned the ChatPainter model on publicly available dialogue data and obtained significant improvement in inception score on the \gls{coco} dataset. While many of the generated $256\times256$ images look quite realistic, the StackGAN family of models (including ChatPainter) has several limitations and also exhibits some of the issues other \glspl{gan} also suffer from. The StackGAN family is able to generate photo-realistic images easily on restricted-domain datasets such as those on flowers and birds but on \gls{coco}, it is able to generate images that exhibit strong global consistency but does not produce recognizable objects in many cases. The current training loss formulation also makes it susceptible to mode collapse. Training the model with dialogue data is also not very stable. Recent improvements in the literature such as training with the WGAN-GP loss can help mitigate these issues to some extent. Using an auxiliary loss for the discriminator by doing object recognition or caption generation from the generated image should also lead to improvements as has been observed in prior work on other image generation tasks. The non-end-to-end training also leads to longer training time and loss of information which can be improved upon by growing the model progressively layer-by-layer as done by \citet{karras2018progressive}.

An interesting research direction we wish to explore further is to generate an image at each turn of the conversation (or modify the previous time-step's image) using dialogues as a feedback mechanism. The datasets we use in this paper neither have separate images for each turn of the dialogue nor is the dialogue dependent on multiple images. In the sketch-artist scenario discussed in Section~\ref{sec:introduction}, the sketch artist would make several changes to the image as the conversation progresses and the future conversation also would depend on the image at that point in the conversation, However, no such publicly available dataset exists yet to the best of our knowledge and we plan to collect such a dataset soon. The recently announced dataset CoDraw \cite{kim2017codraw} contains dialogues about clip art drawings, where intermediate images are updated after every dialogue turn. At the time of publication of this work, CoDraw had not yet been publicly released and if the intermediate images for this dataset are released, that would be a useful contribution for dialogue-to-image-generation research. Image generation guided by dialogue has tremendous potential in the areas of image editing, video games, digital art, accessibility, etc., and is a promising future research direction in our opinion.

\section*{Acknowledgements}
We would like to acknowledge Amjad Almahairi, Kuan-Chieh Wang, and Philip Bachman for helpful discussions on \glspl{gan} and Alex Marino for help with reviewing the generated images. We would also like to thank the authors of StackGAN for releasing their \textit{PyTorch} code which we built upon. This research was enabled in part by support provided by WestGrid and Compute Canada.

\bibliography{example_paper}
\bibliographystyle{icml2018}

\end{document}